\documentclass[10pt,twocolumn,letterpaper]{article}

\newif\ifcvprfinal

\usepackage[pagenumbers]{cvpr}\cvprfinaltrue 

\usepackage{graphicx}
\usepackage{amsmath}
\usepackage{amssymb}
\usepackage{booktabs}

\usepackage{epsfig}
\usepackage{array}
\usepackage{enumitem}
\usepackage{multirow}
\usepackage{caption}
\usepackage{subcaption}
\usepackage{refcount}

\usepackage{dblfloatfix}

\usepackage[pagebackref,breaklinks,colorlinks]{hyperref}

\usepackage[capitalize]{cleveref}
\crefname{section}{Sec.}{Secs.}
\Crefname{section}{Section}{Sections}
\Crefname{table}{Table}{Tables}
\crefname{table}{Tab.}{Tabs.}

\def\cvprPaperID{6835} 
\def\confName{CVPR}
\def\confYear{2023}

\newcommand{\incpageref}[1]{%
  \number\numexpr\getpagerefnumber{#1}}
  
\graphicspath{{figures/}}

\begin{document}

\ifnum8<\incpageref{END_OF_TEXT}
    \title{\textcolor{red}{PAPER IS OVER LENGTH: \incpageref{END_OF_TEXT}}}
\else
    \title{xFBD: Focused Building Damage Dataset and Analysis}
\fi

\author{Dennis Melamed\textsuperscript{1}, Cameron Johnson\textsuperscript{1}, Chen Zhao\textsuperscript{1},\\ Russell Blue\textsuperscript{1}, Philip Morrone\textsuperscript{2}, Anthony Hoogs\textsuperscript{1}, Brian Clipp\textsuperscript{1}\\
\textsuperscript{1}Kitware Inc, \textsuperscript{2}Air Force Research Lab\\
\tt\small \{dennis.melamed, cameron.johnson, chen.zhao, rusty.blue\}@kitware.com,\\ \tt\small philip.morrone.6@us.af.mil,\{anthony.hoogs, brian.clipp\}@kitware.com 
}
\maketitle

\allowdisplaybreaks

\begin{abstract}
   The xView2 competition and xBD dataset \cite{gupta_xbd_2019} spurred significant advancements in overhead building damage detection, but the competition's pixel level scoring can lead to reduced solution performance in areas with tight clusters of buildings or uninformative context. 
   We seek to advance automatic building damage assessment for disaster relief by proposing an auxiliary challenge to the original xView2 competition.
   This new challenge involves a new dataset and metrics indicating solution performance when damage is more local and limited than in xBD.
  Our challenge measures a network's ability to identify individual buildings and their damage level without excessive reliance on the buildings' surroundings.
  Methods that succeed on this challenge will provide more fine-grained, precise damage information than original xView2 solutions.
  The best-performing xView2 networks' performances dropped noticeably in our new limited/local damage detection task. 
  The common causes of failure observed are that (1) building objects and their classifications are not separated well, and (2) when they are, the classification is strongly biased by surrounding buildings and other damage context.
  Thus, we release our augmented version of the dataset with additional object-level scoring metrics (\ifcvprfinal {\small\url{https://drive.google.com/drive/folders/1VuQZuAg6-Yo8r5J4OCx3ZRpa_fv9aaDX?usp=sharing}} \else code \& samples of data available in supplementary for review\fi) to test independence and separability of building objects, alongside the pixel-level performance metrics of the original competition.
  We also experiment with new baseline models which improve independence and separability of building damage predictions.
  Our results indicate that building damage detection is not a fully-solved problem, and we invite others to use and build on our dataset augmentations and metrics.
\end{abstract}

\begin{figure}
    \centering
    \includegraphics[width=\linewidth]{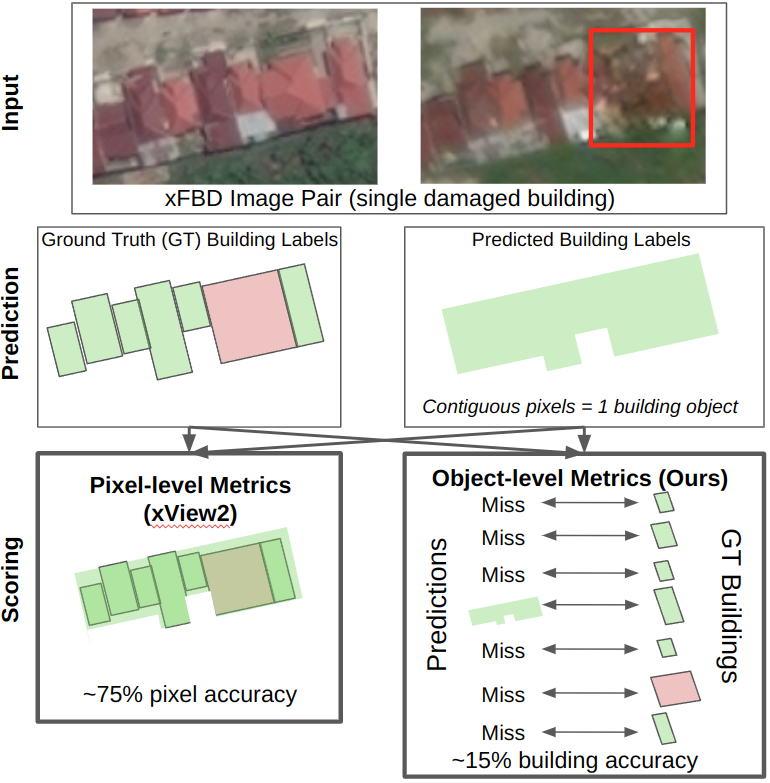}
    \caption{Overview of our proposed xView2 augmentations. On the top is a synthetically created sample from our xFBD dataset with a single damaged building placed in undamaged context. Below, the prediction is scored using different metrics. The xView2 pixel metrics yield a high score, despite the prediction missing the destroyed building and not communicating the presence of 7 buildings. Our object-level metrics provide more insight into performance on this sample. }
    \label{fig:overview_of_contribution}
\end{figure}

\section{Introduction}
When natural disasters occur, humanitarian agencies leverage satellite imagery to assess damage.
Such assessments inform distribution of aid in the hours and days after a disaster. 
Automatic methods to detect buildings and assign damage levels could save hours of work, reducing the time needed to make crisis relief decisions. 
To be dependable, such methods must be robust to variations including unseen geography, weather, and damage distribution. 

The release of the xBD dataset \cite{gupta_xbd_2019} sparked a substantial growth in approaches to automatic building damage assessment.
Our contribution to the growing damage assessment field includes the proposal of new scoring metrics for xBD and augmentations of xBD which we believe will enable more robust building damage assessment.
In particular, we focus on cases where the independence of adjacent building classifications is important.
For example, in earthquakes adjacent buildings can react in markedly different ways due to differences in construction, leading to little correlation between nearby damage levels.  
 
In summary, we find that in xBD adjacent buildings' labels are highly correlated, damage labels often do not relate to the appearance of the building but to its context, and that all of the ``top-five" xView2 solutions are surprisingly blind to damaged buildings in undamaged surroundings.
To quantify this ``context independent classification" capability, we release an augmented version of xBD called xFBD (Focused Building Damage).
Our augmented post-event images contain pre-disaster undamaged content, except for one damaged building which is realistically spliced in.
We introduce some baselines that are compared to the top five xView2 solutions using the original xView2 metrics, alongside new metrics.
These new metrics help quantify network decisions at the building-level instead of the individual pixel scale (see Figure~\ref{fig:overview_of_contribution}).


\section{Description of xView2 challenge}
The xView2 challenge and the related xBD dataset were released in 2019 to advance ``change detection and building damage assessment for humanitarian assistance and disaster recovery research'' \cite{gupta_xbd_2019}.
For 16 disasters of 6 types, pre- and post-disaster geo-registered satellite imagery is provided as RGB chips.
About 850,000 human-labeled building geo-polygons are also provided. 
Each polygon has a damage level based on a four-level scale (Table~\ref{table:damage_scale}).

\begin{figure}
    \centering
    \includegraphics[width=0.85\linewidth]{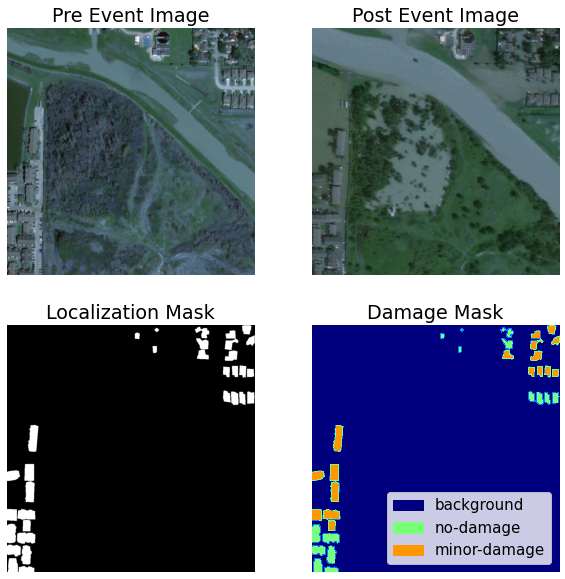}
    \caption{Sample from xBD. A pre- and post-disaster image (top) are given as input to a network, which outputs a localization \& a damage mask (bottom).}
    \label{fig:xbd_sample}
\end{figure}

The xView2 challenge tasked participants to provide a system $f$. Given a pre-disaster image ($x_{pre}$) and a post-disaster image ($x_{post}$), the system should output a building localization mask ($y_{loc}$) and a damage-level prediction mask for pixels predicted to be buildings ($y_{dam}$), i.e. $y_{loc}, y_{dam} = f(x_{pre}, x_{post})$.
Solutions are scored first on the F1 score of their localization prediction, compared pixel-wise against a ground truth mask (Figure~\ref{fig:xbd_sample}, lower left). 
For pixels where a building is predicted, the class-wise damage F1 is also computed pixel-wise against a ground truth mask (Figure~\ref{fig:xbd_sample}, lower right). 
The overall ``score'' for xView2 is a weighted mean of 30\% localization F1 and 70\% overall damage F1. 

\begin{table}
\begin{tabular}{p{4.5em}|p{16em}}
 \textbf{Name}         & \textbf{Description}                                                                                                            \\
\hline
Undamaged    & Undisturbed. No sign of water, structural, or shingle damage, or burn marks                                            \\ \hline
Minor Damage& Partially burnt, water surrounding structure, volcanic flow nearby, roof elements missing, or visible cracks. \\ \hline
Major Damage & Partial collapse, encroaching volcanic flow, or surrounded by water/mud.                                  \\ \hline
 Destroyed    & Scorched, collapsed, covered with water/mud, or no longer present.          
\end{tabular}

\caption{Damage scale used by xView2. Note that not all damage indicators pertain to the actual building pixels, such as ``surrounded by water/mud" indicating ``major damage."}
\label{table:damage_scale}
\end{table}

\section{Related Work}
After the xView2 challenge the top five solutions were released to GitHub \cite{firstplace, secondplace,thirdplace,fourthplace,fifthplace}. 
As can be seen in the first row of Table~\ref{table:object_level_original_xbd}, all of the top results have very close competition scores.
These scores imply the presence of some performance ``ceiling" for this data.
Furthermore, of the 75 published works which have since cited the xBD dataset\cite{gupta_xbd_2019}, there does not appear to be a method which significantly outperforms the top-5 solutions.

The first-place solution \cite{firstplace} implements 4 pre-trained networks re-trained mostly on pre-event xBD images, with occasional post-event images used as augmentation. 
Dice \cite{milletari2016v} and focal \cite{lin2017focal} loss are used to incentivize high-resolution localization, which is important for dense building clusters.

Like the first place solution, the second \cite{secondplace} and third \cite{thirdplace} place solutions use network ensembles. 
The fourth \cite{fourthplace} and fifth \cite{fifthplace} place solutions are much lighter while maintaining close to top level performance. 
Most notable for the work described in this paper, the fourth place solution is described in \cite{zheng2021building}.
This solution uses a siamese network: a pre-event image ResNet \cite{resnet} generates location predictions, and a post-event ResNet fuses pre-event branch features with post-event image features to predict damage.

Since the xView2 competition ended a number of works have used the xBD dataset.  
However, we have not seen marked improvement over the top-5 solutions. 
Da \etal \cite{da2022building} reported a slight improvement in damage F1 scores using a Siamese Hierarchical Transformer Framework.

Bouchard et al. \cite{bouchard2022transfer} notes that the challenge does not capture how well a damage assessment network will adapt to an entirely-unseen city or disaster type.
This uncertainty is due to the train/test data splits being based on different parts of the same general areas of the same disasters.


\section{Need for Further Analysis of xView2}
\label{section:xbd_analysis}
\subsection{Inconsistency of Labelling}
\begin{figure}
    \centering
    \includegraphics[width=0.75\linewidth]{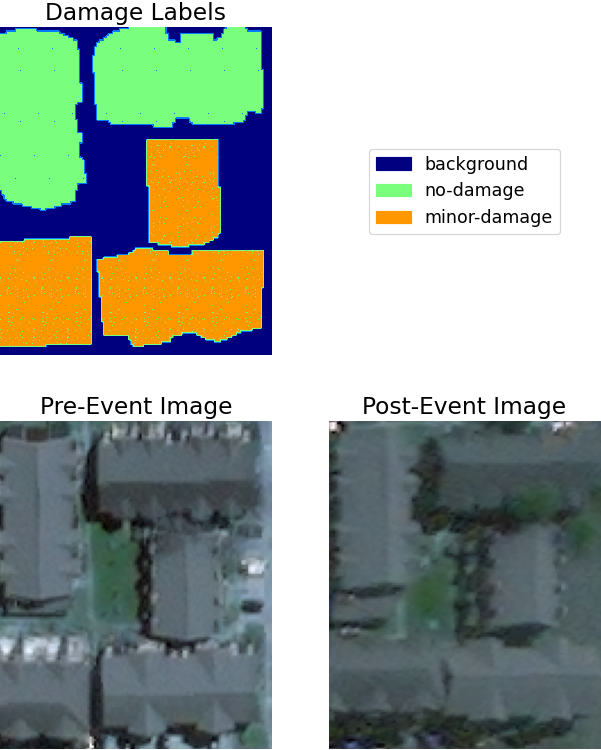}
    \caption{Example of visually difficult-to-distinguish damage levels. The lower buildings show no obvious damage, yet are labeled as minor damage in contrast to the undamaged upper buildings.}
    \label{fig:labeling_interp_example}
\end{figure}

The xView2 challenge can be considered to be a 5-class pixel-wise classification problem. 
Every location in an image must be classified as one of: not-building (background), no-damage, minor-damage, major-damage, or destroyed. 
Upon visual inspection, it is not clear that the 4 damage categories are always distinguishable by any visible feature on the roof. 
Figure~\ref{fig:labeling_interp_example} shows an example of such a case.
The features of each building do not provide obvious evidence that these buildings belong to different classes.
This class ambiguity is supported by the challenge's descriptions of damage classes (Table \ref{table:damage_scale}).
These descriptions do not all pertain to pixels belonging to a building (building pixels) but frequently to the building's context.
The lack of distinguishing features between classes may lead to network confusion and lower performance across categories.

In order to further analyze the xView2 challenge solutions, we select the fourth place solution (which we call \textbf{4PS}) developed by Zheng et al. \cite{zheng2021building} as an exemplar. 
The 4PS was selected to avoid lengthy model ensemble training/inference time while maintaining top xView2 performance.
All top 5 solutions' overall scores are within 0.03, so it is likely they will perform similarly in our experiments.
We re-implemented the 4PS in Pytorch-Lightning \cite{Falcon_PyTorch_Lightning_2019}.

Using the 4PS, we can analyze if the four damage categories contain enough signal for networks to usefully distinguish them.
We collapse the classification problem into two classes (low-/high-damage) from the original four.
The 4PS has relatively low performance on the minor and major damage classes of the original problem, but when predictions are collapsed into two categories the overall performance improves (Table~\ref{tab:binary_problem}). 
The network is able to distinguish between mostly untouched vs. damaged buildings, but has more difficulty making decisions in the 4 class space.
The four damage categories may be too subtly defined for a network to effectively distinguish them.

\begin{table}

\begin{tabular}{l|l|l}
             & Pixel Level F1 Scores & Sample Count \\
        \hline
No Damage    & 0.9264                & 313,033      \\
Minor damage & 0.5969                & 38,860       \\
Major Damage & 0.7442                & 29,904       \\
Destroyed    & 0.8317                & 31,560      
\end{tabular}

\begin{tabular}{l|l}
                         & Pixel Level F1 Scores \\
                         \hline
No Damage + Minor Damage & 0.9377                \\
Major Damage + Destroyed & 0.8234            

\end{tabular}
\caption{
Damage F1 scores are significantly improved when ``no damage/minor damage" and ``major damage/destroyed" confusion are eliminated.
}
\label{tab:binary_problem}
\end{table}

\subsection{Object-level Performance}

The xView2 challenge focused on pixel-level F1 scores. 
However, it may be more intuitive to understand damage at a building (instead of pixel) level. 
For example, we may wish to understand how many homes have been destroyed to determine relief requirements, instead of the ``destroyed area'' estimate that a pixel-level paradigm produces. 
Thus, we propose a set of object-level metrics to quantify how networks understand damage on a building level. 

The object-level metrics are formulated using the xBD label files containing building polygons associated with labels. 
These are converted to a COCO object-detection dataset \cite{coco}. 
A similar COCO dataset is built from a solution's localization and damage prediction masks. 
The localization mask is converted into objects using connected component labeling. 
For each predicted building, the damage mask pixels belonging to it determine the its label via majority vote. 
Object F1 scores are calculated via KWCOCO \cite{kwcoco} at IoU=0.5 for overall localization and the 4 damage categories. 

\begin{figure}
     \centering
     \begin{subfigure}[b]{0.49\linewidth}
         \centering
         \includegraphics[width=0.8\linewidth]{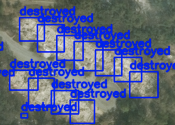}
         \caption{Ground truth bounding boxes}
         \label{fig:building_swath_example_gt}
     \end{subfigure}
     \hfill
     \begin{subfigure}[b]{0.49\linewidth}
         \centering
         \includegraphics[width=0.8\linewidth]{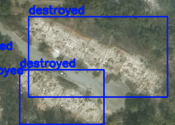}
         \caption{Predicted bounding boxes}
         \label{fig:building_swath_example_pred}
     \end{subfigure}

    \caption{Example of the 4PS predicting a ``swath'' of damage, fusing multiple buildings together. Using our object-level metrics, this network fails for all but 2 of the buildings.}
    \label{fig:building_swath_example}
\end{figure}

Table~\ref{table:object_level_original_xbd} shows the top-5 solutions' xBD holdout set performance on our object-level metrics. 
The original xView2 overall score is provided to verify that our codebases have similar performance to the originally submitted solutions.
Performance in all object-level scores is fairly low.
While these numbers cannot be directly compared to the pixel-wise scores, as they measure very different things, the low object-level scores indicate that top-5 networks do not tend to make per-building decisions.
Instead, they group and predict labels for `swathes' of buildings together (see Figure~\ref{fig:building_swath_example}).
In the xView2 challenge, the network is not penalized for making one prediction for a group of several buildings if they all share a damage label, leading to the formation of connected swathes of buildings. 

\begin{table}
\small
\begin{tabular}{llllll}
                & \multicolumn{5}{c}{xView2 Solution \#}    \\ 
\multicolumn{1}{l|}{}                   & \multicolumn{1}{l|}{\textbf{1}} & \multicolumn{1}{l|}{\textbf{2}} & \multicolumn{1}{l|}{\textbf{3}} & \multicolumn{1}{l|}{\textbf{4}} & \multicolumn{1}{l|}{\textbf{5}} \\ \cline{2-6} 
\multicolumn{1}{l|}{\begin{tabular}[c]{@{}l@{}}xView2 Comp. Score \end{tabular}} & \multicolumn{1}{l|}{0.78}  & \multicolumn{1}{l|}{0.81}  & \multicolumn{1}{l|}{0.79}  & \multicolumn{1}{l|}{0.76}  & \multicolumn{1}{l|}{0.73}  \\ \hline
\multicolumn{1}{l|}{\begin{tabular}[c]{@{}l@{}}Object Localization F1\end{tabular}}                              & \multicolumn{1}{l|}{0.44}  & \multicolumn{1}{l|}{0.43}  & \multicolumn{1}{l|}{0.42}  & \multicolumn{1}{l|}{0.40}  & \multicolumn{1}{l|}{0.45}  \\ \hline
\multicolumn{1}{l|}{\begin{tabular}[c]{@{}l@{}}No Damage Object F1\end{tabular}}                                    & \multicolumn{1}{l|}{0.45}  & \multicolumn{1}{l|}{0.41}  & \multicolumn{1}{l|}{0.40}  & \multicolumn{1}{l|}{0.36}  & \multicolumn{1}{l|}{0.41}  \\ \hline
\multicolumn{1}{l|}{\begin{tabular}[c]{@{}l@{}}Minor Damage Object F1\end{tabular}}                                 & \multicolumn{1}{l|}{0.16}  & \multicolumn{1}{l|}{0.25}  & \multicolumn{1}{l|}{0.21}  & \multicolumn{1}{l|}{0.24}  & \multicolumn{1}{l|}{0.21}  \\ \hline
\multicolumn{1}{l|}{\begin{tabular}[c]{@{}l@{}}Major Damage Object F1\end{tabular}}                                 & \multicolumn{1}{l|}{0.24}  & \multicolumn{1}{l|}{0.27}  & \multicolumn{1}{l|}{0.25}  & \multicolumn{1}{l|}{0.30}  & \multicolumn{1}{l|}{0.32}  \\ \hline
\multicolumn{1}{l|}{\begin{tabular}[c]{@{}l@{}}Destroyed Object F1\end{tabular}}                                    & \multicolumn{1}{l|}{0.33}  & \multicolumn{1}{l|}{0.38}  & \multicolumn{1}{l|}{0.37}  & \multicolumn{1}{l|}{0.37}  & \multicolumn{1}{l|}{0.40}  \\ 
\end{tabular}

\caption{Object-level xView2 solution scores, xBD holdout set.}
\label{table:object_level_original_xbd}
\end{table}

\subsection{Influence of Context on Network Performance}

\noindent\textbf{Experimental Setup.}
From our understanding that xView2 solutions tend to predict swathes of damage, as opposed to making building-wise decisions, the question of the importance of context arises. 
Are solutions making damage decisions based on building features or on the appearance of the surrounding area? 
For disasters such as flooding, this is understandable.
Buildings may be left standing but the surrounding water makes it clear the structure is destroyed. 
However, for other disasters this prior may cause unintended consequences.
A tornado may rip through a town, but a building next to the destruction may be spared. 
If xView2 solutions have a heavy dependence on context and are used to determine disaster relief allocations, resources are likely to be sent to areas with widespread damage. 
But, single damaged buildings surrounded by undamaged buildings might be missed and not receive needed aid. 
Thus, we seek to quantify context's importance to xView2 solutions. 

\begin{figure}
    \centering
    \includegraphics[width=0.8\linewidth]{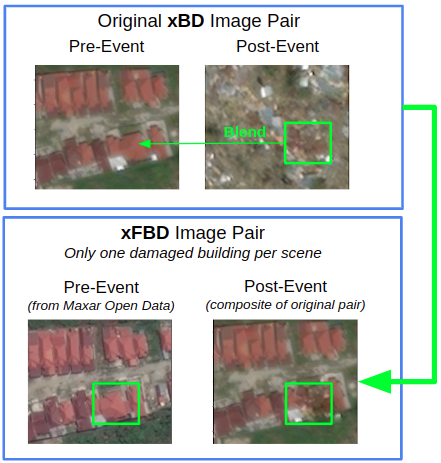}
    \caption{Example of a single post-event building blended into pre-event context. This creates an image pair with no damage evidence except the one ``post-disaster" building.}
    \label{fig:single_building_paste_ex}
\end{figure}

In order to test the influence of context we form a composite post-disaster image, containing one damaged building from the original post-disaster image.
All other buildings, along with the background, are drawn from the pre-disaster image. 
An example of this blending procedure is shown in Figure~\ref{fig:single_building_paste_ex}.
Blending in a single building tests whether solutions can detect hyper-local damage. 


Our composite image is formed via Poisson Blending \cite{poisson_blending} to enable a realistic blend.
The original xBD labels are used to form a mask of which semantic area (i.e. the damaged building of interest) of the post-image to blend into the pre-image. 
The target network output masks are updated accordingly, with all buildings re-labeled as ``undamaged" except for the blended building which retains its label.
In generating our augmented data, we seek to avoid creating a very simple change detection problem where the only pixel-level change between pre- and post-disaster is the blended building.
Thus, a secondary pre-disaster image is downloaded from Maxar's Open Data Program \cite{maxar_open_data} from a different date than the original.
The Palu Tsunami disaster is chosen for this experiment as it has a balanced mixture of damage levels in both damaged and undamaged contexts, and the secondary pre-images chosen for it are high-quality and well geo-aligned with the originals. 

To verify that these secondary pre-images are appropriate for use we predict on the holdout set using the 4PS, giving the network the secondary pre-image as its pre-event input and the original post-image as its post-event input. 
The results in Table~\ref{table:secondary_pre_image_verification} indicate that performance does not drop significantly with the new pre-images, which gives us confidence that we have not missed some artifact in the new imagery that could affect results. 
Further experiments all test using the new pre-images.
Therefore, comparisons between methods which are trained on the original pre-images remain fair, since all methods would experience the relatively small drop in performance from using the new pre-image.


\begin{table}[]
    \centering

\begin{tabular}{ll|l|l}
                                                                                  &      \hspace{30pt}\textbf{Pre-Image:}        & \begin{tabular}[c]{@{}l@{}}Original\\ \end{tabular} & \begin{tabular}[c]{@{}l@{}}Secondary \\ \end{tabular} \\ \cline{2-4} 
\multirow{5}{*}{\begin{tabular}[c]{@{}l@{}}Pixel\\ Level\\ Scores\end{tabular}}   & Localization F1 & 0.8953                                                               & 0.7812                                                                 \\ \cline{2-4} 
                                                                                  & No Damage F1    & 0.9578                                                               & 0.9056                                                                 \\ \cline{2-4} 
                                                                                  & Minor Damage F1 & \begin{tabular}[c]{@{}l@{}}n/a \end{tabular}          & \begin{tabular}[c]{@{}l@{}}n/a \end{tabular}            \\ \cline{2-4} 
                                                                                  & Major Damage F1 & 0.1845                                                               & 0.2169                                                                 \\ \cline{2-4} 
                                                                                  & Destroyed F1    & 0.7701                                                               & 0.6968                                                                 \\ \hline \hline
\multirow{5}{*}{\begin{tabular}[c]{@{}l@{}}Object\\ Level \\ Scores\end{tabular}} & Localization F1 & 0.13                                                                 & 0.06                                                                   \\ \cline{2-4} 
                                                                                  & No Damage F1    & 0.12                                                                 & 0.05                                                                   \\ \cline{2-4} 
                                                                                  & Minor Damage F1 & \begin{tabular}[c]{@{}l@{}}n/a \end{tabular}          & \begin{tabular}[c]{@{}l@{}}n/a \end{tabular}            \\ \cline{2-4} 
                                                                                  & Major Damage F1 & 0.02                                                                 & 0.01                                                                   \\ \cline{2-4} 
                                                                                  & Destroyed F1    & 0.16                                                                 & 0.07                                                                   \\ 
\end{tabular}

    \caption{4PS results on original xBD vs. our secondary pre-event images (holdout set). No secondary pre-event images for samples with minor damage were available.}
    \label{table:secondary_pre_image_verification}
\end{table}


\begin{figure*}
    \centering
    \includegraphics[width=0.8\linewidth]{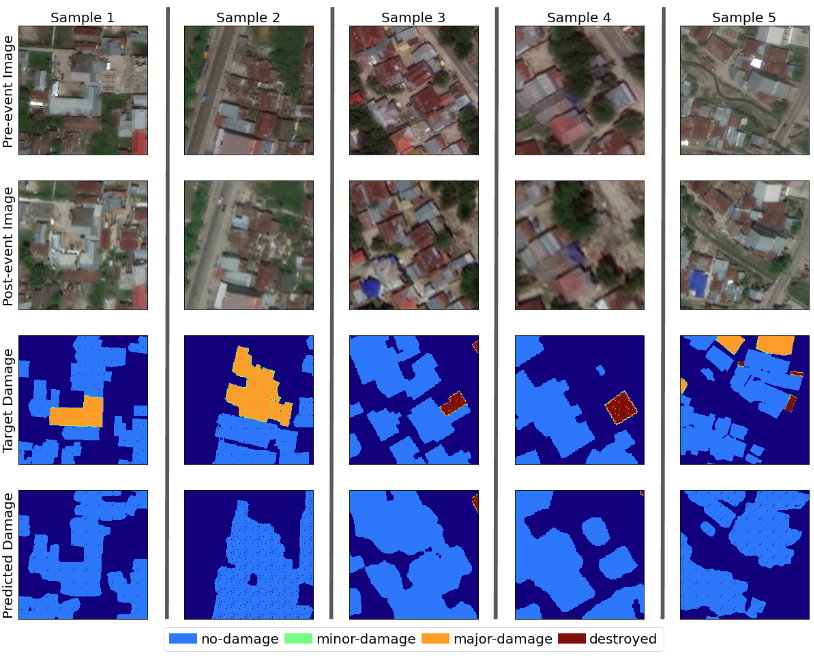}
    \caption{Localized damage examples from original xBD dataset. A single or small cluster of buildings is damaged, but the surroundings are undamaged. The 4PS network assumes that the building's class is the same as surrounding buildings', despite obvious damage.}
    \label{fig:xbd_isolated_buildings}
\end{figure*}

\noindent\textbf{Results.}
4PS pixel and object-level results are shown in Table~\ref{table:top_5_xfbd_results}, column \textbf{4}.
Damage F1 scores for buildings blended in are low, indicating that the network is not triggering on the damage features of a given building.
Instead, it is likely classifying based on surrounding buildings/context.



\begin{figure}
    \centering
    \includegraphics[width=0.85\linewidth]{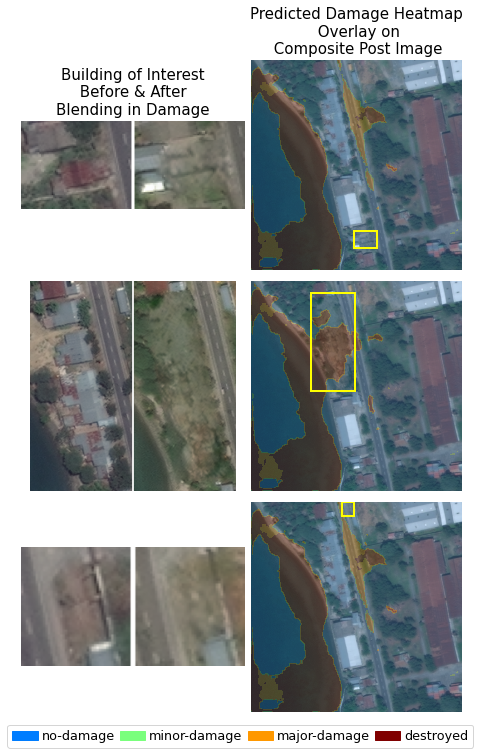}
    \caption{Right: The 4PS damage heatmap for single-damaged-building samples is overlaid on the composite post-image. As different buildings are blended, the heatmap only changes significantly for large buildings. Left: a zoom of the target building.}
    
    \label{fig:blend_unchanging_heatmap}
\end{figure}

Further qualitative understanding can be gained from analyzing the underlying damage heatmap output by the 4PS.
The heatmap is the final stage of the 4PS damage branch, which is later combined with the predicted building locations to generate the final output. 
In Figure~\ref{fig:blend_unchanging_heatmap}, different damaged buildings (left) are blended into the same undamaged pre-event image.
The heatmap output from the composite images is shown on the right. 
Of interest is the relatively minor changes in the damage heatmap, even as the building which is damaged changes.
The only significant change in the heatmap occurs when the largest building is replaced with its damaged post-event counterpart. 
Small buildings being damaged appear to not affect the network's output, further evidence that the network is not classifying on a building-level basis. 
Instead, wide areas are used to make more global decisions for all buildings in particular areas. 
Further evidence of this issue is shown in Figure~\ref{fig:xbd_isolated_buildings}, which shows several examples of damaged buildings surrounded by undamaged buildings in the original xBD dataset.
The 4PS network is unable to correctly classify these samples, instead lumping them in with nearby buildings.

\section{xFBD dataset \& challenge formulation }


In order to further research in this area, we release the dataset created for the single damaged building blend experiment.
We also formulate an auxiliary challenge to xView2 called xFBD: Focused Building Damage Assessment. 
xFBD is not intended to replace xView2, which provides valuable damage prediction performance information. 
xFBD instead supplements the understanding of a solution's performance in two areas: \textbf{1}) The ability to distinguish buildings from each other and thus provide more accurate building counts, and \textbf{2}) The ability to make classifications from individual buildings' features instead of wider areas of damage, increasing confidence in individual building assessments.
Table~\ref{table:xfbd_dataset_stats} shows descriptive statistics of the xFBD dataset.
Not all Palu Tsunami images in xBD are used to generate samples as some do not have a high quality secondary pre-event image available. 

Solutions are scored on both the original xBD dataset and the new xFBD dataset.
Better performance on xFBD, ideally, would not cause a drop in xBD performance.
Each solution is scored on both the original pixel-level xView2 metrics and the object-level metrics introduced here. 

A connected component analysis was used in our experiments with the top xView2 solutions to generate object-level predictions from the predicted semantic masks.
We do not require this method be used: solutions may generate object- and pixel-level predictions in parallel or in any order.
For example, Mask-RCNN \cite{maskrcnn} first generates an object-level prediction, then a pixel mask for the object.

\begin{table}[]
    \centering
\begin{tabular}{llll}
\multicolumn{1}{l|}{}                                                                            & \multicolumn{1}{l|}{train (tier1)} & \multicolumn{1}{l|}{test} & hold \\ \hline
\multicolumn{1}{l|}{\begin{tabular}[c]{@{}l@{}}\# samples (image pairs)\end{tabular}}              & \multicolumn{1}{l|}{5287}          & \multicolumn{1}{l|}{1466} & 1274 \\ 
\multicolumn{1}{l|}{\# source images used}                                                            & \multicolumn{1}{l|}{76}            & \multicolumn{1}{l|}{27}   & 28   \\ 
\multicolumn{1}{l|}{\begin{tabular}[c]{@{}l@{}}\# destroyed samples \end{tabular}}       & \multicolumn{1}{l|}{4747}          & \multicolumn{1}{l|}{1235} & 940  \\ 
\multicolumn{1}{l|}{\begin{tabular}[c]{@{}l@{}}\# major damage samples\end{tabular}} & \multicolumn{1}{l|}{539}           & \multicolumn{1}{l|}{231}  & 334  \\ 
\multicolumn{1}{l|}{\begin{tabular}[c]{@{}l@{}}\# minor damage samples\end{tabular}} & \multicolumn{1}{l|}{1}             & \multicolumn{1}{l|}{0}    & 0   
\end{tabular}
    \caption{Descriptive statistics of the xFBD dataset.}
    \label{table:xfbd_dataset_stats}
\end{table}

\subsection{xFBD Results for Current SOTA}

In comparison to the original xView2 results, the top-5 solutions perform poorly on the xFBD dataset (Table \ref{table:top_5_xfbd_results}). 
In the case of a single destroyed building in an otherwise-undamaged landscape one might expect still-reasonable performance, but the best object-level F1 from the top-5 is 0.06. 
This indicates that the top networks are more dependent on adjacent buildings and other damage context than is useful for hyper-local damage detection.

\begin{figure}
    \centering
    \includegraphics[width=0.6\linewidth]{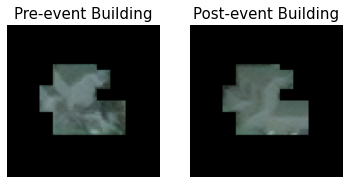}
    \caption{Example of classifier input: a single building chip pair. Only pixels from one building are used for inference.}
    \label{fig:single_building_chip_ex}
\end{figure}

\begin{table}
    \centering

\begin{tabular}{ll|wc{1.25em}|wc{1.25em}|wc{1.25em}|wc{1.25em}|wc{1.25em}|}
                                                                                  &                                                           & \textbf{ 1} & \textbf{2} & \textbf{3} & \textbf{4} & \textbf{5} \\ \cline{3-7} 
\parbox[t]{2mm}{\multirow{5}{*}{\rotatebox[origin=c]{90}{\textbf{Pixel-Level}}}}
& \begin{tabular}[c]{@{}l@{}}Localization F1\end{tabular}  & 0.76  & 0.77  & 0.76  & 0.76  & 0.75  \\ \cline{2-7} 
                                                                                  & \begin{tabular}[c]{@{}l@{}}No  Damage F1\end{tabular}    & 0.88  & 0.91 & 0.91  & 0.91  &  0.85 \\ \cline{2-7} 
                                                                                  & \begin{tabular}[c]{@{}l@{}}Minor  Damage F1\end{tabular} & n/a  &  n/a & n/a & n/a &  n/a \\ \cline{2-7} 
                                                                                  & \begin{tabular}[c]{@{}l@{}}Major  Damage F1\end{tabular} & 0.00  & 0.01  & 0.01  & 0.02  & 0.01  \\ \cline{2-7} 
                                                                                  & \begin{tabular}[c]{@{}l@{}}Destroyed  F1\end{tabular}    & 0.08  & 0.11 & 0.07  & 0.07  &  0.07 \\ \hline\hline
\parbox[t]{2mm}{\multirow{5}{*}{\rotatebox[origin=c]{90}{\textbf{Object-Level}}}}
& \begin{tabular}[c]{@{}l@{}}Localization F1\end{tabular} & 0.06  & 0.06  & 0.05  & 0.04  & 0.06  \\ \cline{2-7} 
                                                                                  & \begin{tabular}[c]{@{}l@{}}No Damage F1\end{tabular}    & 0.06  & 0.06 & 0.05  &  0.04 &  0.06 \\ \cline{2-7} 
                                                                                  & \begin{tabular}[c]{@{}l@{}}Minor Damage F1\end{tabular} &  n/a & n/a & n/a  & n/a  &  n/a \\ \cline{2-7} 
                                                                                  & \begin{tabular}[c]{@{}l@{}}Major Damage F1\end{tabular} & 0.00  & 0.00 & 0.00  & 0.00  &  0.00 \\ \cline{2-7} 
                                                                                  & \begin{tabular}[c]{@{}l@{}}Destroyed  F1\end{tabular}    & 0.00  & 0.00  & 0.01  & 0.00  &  0.00 \\ 
\end{tabular}

    \caption{Results of running xView2 solutions on the xFBD dataset. All of the top solutions lack the ability to identify damaged buildings when surrounding context implies ``no-damage"}
    \label{table:top_5_xfbd_results}
\end{table}

\subsection{Realism of xFBD Images}
We must ensure that xFBD images are realistic enough to be useful training and evaluation data.  
To do so, we use xFBD samples to finetune the 4PS network.
We examine the finetuned performance on a test set of real isolated damaged buildings drawn from original xBD data. 

Our test set is generated by first filtering all xBD building labels present in the xFBD test and hold sets.
Major-damage or destroyed buildings were selected which did not have other damaged buildings within a 50 pixel radius. 
These buildings were then hand-filtered to ensure they represented true isolated damage: samples where the building was surrounded by obvious evidence of destruction such as destroyed forest were removed. 
This filtering left 45 damaged samples. 
45 undamaged buildings were also selected from xFBD to balance the test set. 

A 4PS network trained on xBD was finetuned using the xFBD train split.
We then evaluated both the original and finetuned version of the 4PS on our real isolated damage test images. 
Scoring was done only on pixels belonging to the 90 buildings from our test set. 
Results are shown in Table~\ref{tab:real_isolated_damage_results}. 

\begin{table}
\begin{tabular}{l|l|l}
                     & Original & Finetuned xFBD \\
                     \hline
    No-damage F1 & 0.696 & 0.661 \\
    Minor-damage F1 & n/a & n/a \\
    Major-damage F1 & 0.000 & 0.508 \\
    Destroyed F1 & 0.000 & 0.192

\end{tabular}
\caption{4PS performance on detecting real examples of damaged buildings with undamaged context, with (Finetuned xFBD) and without (Original) finetuning on synthetic xFBD samples.}
\label{tab:real_isolated_damage_results}
\end{table}

The original 4PS network fails to detect the isolated damaged buildings in our test set. 
However, by introducing xFBD samples during training the network begins to detect such situations in real life. 
These results suggest that our synthetic samples are realistic enough to be a good proxy for real buildings where damage has occurred but damage to the building's surroundings has not. 

\subsection{Tight Building Chip Experiment}
We wish to verify that there is enough information present in the dataset we provide to get better object-level results than the current top-5 solutions.
Thus, we use EfficientNet-b7 \cite{efficientnet} as a simple classifier to show that with ideal localization performance, it is possible to improve on  current best object-level damage F1 scores. 
An example input to the EfficientNet is shown in Figure~\ref{fig:single_building_chip_ex}.
Individual buildings are chipped from the images available in the xBD training set, and are cropped to the xBD annotated building polygon. 
Both the pre-event and post-event chips are combined to form a six channel image, which is then padded to create a consistent input size. 
The network is asked to classify the building chip as one of the four damage classes. 
This is a relatively unoptimized experiment but is able to outperform xView2 solutions on object-level metrics for the holdout set of buildings, as seen in Table \ref{table:single_building_chip_results}. 
We compare against the 5th place xView2 solution as a representative of the performance level of xView2 solutions overall. 
Thus, we feel there is more object-level performance a network could achieve on xFBD by just looking at detected building pixels and not relying so heavily on context.

\begin{table}
    \centering
    \begin{tabular}{p{7.25em}|p{7em}|p{5em}}
         Object Level~Results & \textbf{Chip Classifier} & \textbf{xView2~5th Place~Model}\\
         \hline
         Localization F1 & n/a (pre-chipped) & 0.45 \\
         No-damage F1 &  0.68 & 0.41\\
         Minor-damage F1 & 0.46 & 0.21\\
         Major-damage F1 &  0.47 & 0.32\\
         Destroyed F1 &  0.36 & 0.40
    \end{tabular}
    \caption{Performance of the tight building chip classifier on the original xBD dataset. Inference using just the pixels on the roof of a building can yield reasonable performance.}
    \label{table:single_building_chip_results}
\end{table}


\section{New Baseline Models}

\begin{table*}[h]
\begin{subtable}[h]{\linewidth}
\centering
        \begin{tabular}{l|l|ccccc}
            \hline
             & Method & Localization & Destroyed & Major Damage & Minor Damage & No Damage \\
            \hline
            \parbox[t]{2mm}{\multirow{4}{*}{\rotatebox[origin=c]{90}{Pix-Level}}} & 4PS & \textbf{0.860} & \textbf{0.822} & \textbf{0.719} & 0.508 & \textbf{0.922} \\
            & 4PS-Focal & 0.857 & 0.794 & 0.694 & 0.513 & 0.907 \\
            & 4PS-Contour & 0.849 & 0.787 & 0.703 & \textbf{0.569} & 0.908\\
            & Mask-RCNN &0.699 & 0.605 & 0.429& 0.277&0.675 \\
            \hline
            \parbox[t]{2mm}{\multirow{4}{*}{\rotatebox[origin=c]{90}{Obj-Level}}} & 4PS &  0.440 & 0.420 & 0.340 & 0.240 & 0.420 \\
            & 4PS-Focal & 0.440 & \textbf{0.430} & \textbf{0.360} & 0.210 & 0.400 \\
            & 4PS-Contour & 0.410 & 0.350 & 0.310 & \textbf{0.240} & 0.360\\
            & Mask-RCNN & \textbf{0.470} & 0.280 & 0.300 &0.180 & \textbf{0.450} \\
            \hline
        \end{tabular}
    \caption{Original xBD holdout set}
    \label{table:xbd_baselines}
\end{subtable}

\begin{subtable}[h]{\linewidth}
\centering
\begin{tabular}{l|l|ccccc}
            \hline
             & Method & Localization & Destroyed & Major Damage & Minor Damage & No Damage \\
            \hline
            \parbox[t]{2mm}{\multirow{4}{*}{\rotatebox[origin=c]{90}{Pix-Level}}} & 4PS & \textbf{0.758} & \textbf{0.066} & \textbf{0.016} & n/a & 0.907  \\
            & 4PS-Focal & 0.742 & 0.042 & 0.000 & n/a &  0.873\\
            & 4PS-Contour & 0.741 & 0.060 & 0.000 & n/a & 0.887\\
            & Mask-RCNN & 0.494 & 0.013& 0.011& n/a & 0.522\\
            \hline
            \parbox[t]{2mm}{\multirow{4}{*}{\rotatebox[origin=c]{90}{Obj-Level}}} & 4PS & 0.040 & 0.000 & 0.000 & n/a & 0.040  \\
            & 4PS-Focal & 0.040 & \textbf{0.010} & 0.000 & n/a & 0.010  \\
            & 4PS-Contour & 0.050 & 0.000 & 0.000 & n/a & 0.050\\
            & Mask-RCNN & \textbf{0.210} & 0.000 & 0.000 & n/a & \textbf{0.200} \\
            \hline
        \end{tabular}

    \caption{Single-building-blend dataset (xFBD)}
    \label{table:xfbd_baselines}
    \end{subtable}
        \vspace{-10px}
    \caption{Results of New Baselines Proposed}
\end{table*}
We develop new xFBD baselines by refactoring the 4PS loss function and exploring the original xView2 baseline.
Tables \ref{table:xbd_baselines} and \ref{table:xfbd_baselines} show these methods are able to outperform the original 4PS on our object-level metrics, but more work is needed to achieve significant improvements. 
The following losses and models are proposed to improve damage prediction by better separating connected objects:

    
    \textbf{4PS-Focal}. Extended from binary cross entropy loss, focal loss \cite{lin2017focal} was initially designed to improve one-stage object detection. 
    We train a soft dice (SD) loss alongside a version of the focal loss $L_R$ where a modulating factor is added to the cross entropy loss with tuneable focusing parameter $\gamma\geq 0$. The 4PS-Focal loss is:
    \begin{gather}
        L_{Focal}(\mathbf{y},\mathbf{\hat{y}}) = L_{SD}(\mathbf{y},\mathbf{\hat{y}}) + L_{R}(\mathbf{y},\mathbf{\hat{y}})\\
        L_{SD}(\mathbf{y},\mathbf{\hat{y}}) = \frac{1-\mathbf{2}^T(\mathbf{y}\odot\mathbf{\hat{y}})}{\mathbf{1}^T(\mathbf{y}+\mathbf{\hat{y}})}\\
        L_{R}(\mathbf{y},\mathbf{\hat{y}}) 
        = -(1- \sum_i^n h(y_i,\hat{y}_i))^{\gamma}\log(\sum_i^n h(y_i,\hat{y}_i)) \label{eq:focal}\\
        h(y_i,\hat{y}_i) = \sum_i^n (1-y_i)(1-\hat{y}_i) + y_i\hat{y}_i
    \end{gather}
    $n$ is the total number of pixels in an image, and $\odot$ is element-wise product. 
    $\gamma$ controls the effect of the modulating factor. We found $\gamma=2$ to work best in our experiments.

     \textbf{4PS-Contour}. A contour loss \cite{deng2018learning, fang2021coarse} is introduced to help the network learn more accurate (and separable) boundaries for buildings. 
     The contour loss function is:
    \begin{align}
        L_{Contour}(\mathbf{y},\mathbf{\hat{y}}) = \lambda\cdot L_{cls}(\mathbf{y},\mathbf{\hat{y}})+ L_{dice}(\mathbf{y},\mathbf{\hat{y}})\\
        L_{dice}(\mathbf{y},\mathbf{\hat{y}}) = 1-\frac{\mathbf{1}^T (\mathbf{y}\odot\mathbf{\hat{y}})+\epsilon}{\mathbf{1}^T(\mathbf{y}\odot\mathbf{y})+\mathbf{1}^T(\mathbf{\hat{y}}\odot\mathbf{\hat{y}})+\epsilon}
        \label{eq:dice}
    \end{align}
    where
    $L_{cls}(\mathbf{y},\mathbf{\hat{y}})=-\sum_i^n\big(y_i\log\hat{y}_i+(1-y_i)(1-\log\hat{y}_i)\big)$. 
    $L_{dice}$ is a pixel loss \cite{deng2018learning}, describing the similarity of two sets of image pixels. 
    To avoid a zero denominator, $\epsilon$ is 1.

 \textbf{Mask-RCNN} \cite{maskrcnn} is the original baseline model for xView2.
    While this model did not yield good results on pixel-level metrics, it has a built-in understanding of objects as an object-detection network. 
    It is thus able to achieve the best object-level xBD localization and performs significantly better than 4PS-based models in xFBD object-level localization.
    Further work is needed to improve object classification and pixel scores, but Mask-RCNN shows promise in resolving some issues present in xView2 solutions.%

\section{Conclusion}
We have demonstrated that xView2 solution networks are highly dependent on a building's surrounding context for performance.
Isolated damaged buildings present a considerable challenge.
In scenarios where accurate localization of individual damaged buildings is important, incentivizing networks to independent decisions for separate building objects could help disaster analysts significantly.
The augmented dataset we propose features pre/post event image pairs with only one building changing damage state between the two images, through a Poisson Blend process.

Future versions augmenting the entire xBD dataset for all disasters may enable an increase in ``independent building analysis" performance, as measured by a ``single-building paste" experiment (Table \ref{table:top_5_xfbd_results}). 
A filter must be applied to exclude buildings from the blending process whose damage is not visible from roof pixels, like many buildings in flooding disasters.
The current baseline methods have room to grow for the new object-level metrics proposed. 
More saliency-related research is also needed to understand any artifacts of the augmentation and how those may influence network localization and classification decisions. 

By developing object-level damage metrics and a dataset focused on isolated building damage detection, our paper helps to advance artificial intelligence for disaster relief.

\ifcvprfinal \textbf{Acknowledgements:} This material is based upon work supported by Air Force Research Laboratory, Analytical Systems Branch  (AFRL/RIED) under contract number FA8750-21-C-1514. \fi

\label{END_OF_TEXT}
{\small
\bibliographystyle{ieee_fullname}
\bibliography{egbib}

\begin{thebibliography}{10}\itemsep=-1pt

\bibitem{bouchard2022transfer}
Isabelle Bouchard, Marie-{\`E}ve Rancourt, Daniel Aloise, and Freddie
  Kalaitzis.
\newblock On transfer learning for building damage assessment from satellite
  imagery in emergency contexts.
\newblock {\em Remote Sensing}, 14(11):2532, 2022.

\bibitem{da2022building}
Yifan Da, Zhiyuan Ji, and Yongsheng Zhou.
\newblock Building damage assessment based on siamese hierarchical transformer
  framework.
\newblock {\em Mathematics}, 10(11):1898, 2022.

\bibitem{deng2018learning}
Ruoxi Deng, Chunhua Shen, Shengjun Liu, Huibing Wang, and Xinru Liu.
\newblock Learning to predict crisp boundaries.
\newblock In {\em Proceedings of the European Conference on Computer Vision
  (ECCV)}, pages 562--578, 2018.

\bibitem{firstplace}
Victor Durnov.
\newblock xview2\_first\_place.
\newblock \url{https://github.com/DIUx-xView/xView2_first_place}, 2020.

\bibitem{Falcon_PyTorch_Lightning_2019}
William Falcon and {The PyTorch Lightning team}.
\newblock {PyTorch Lightning}, 3 2019.

\bibitem{fang2021coarse}
Fang Fang, Kaishun Wu, Yuanyuan Liu, Shengwen Li, Bo Wan, Yanling Chen, and
  Daoyuan Zheng.
\newblock A coarse-to-fine contour optimization network for extracting building
  instances from high-resolution remote sensing imagery.
\newblock {\em Remote Sensing}, 13(19):3814, 2021.

\bibitem{gupta_xbd_2019}
Ritwik Gupta, Bryce Goodman, Nirav Patel, Ricky Hosfelt, Sandra Sajeev, Eric
  Heim, Jigar Doshi, Keane Lucas, Howie Choset, and Matthew Gaston.
\newblock Creating xbd: A dataset for assessing building damage from satellite
  imagery.
\newblock In {\em Proceedings of the IEEE/CVF Conference on Computer Vision and
  Pattern Recognition (CVPR) Workshops}, June 2019.

\bibitem{maskrcnn}
Kaiming He, Georgia Gkioxari, Piotr Dollár, and Ross Girshick.
\newblock Mask r-cnn.
\newblock In {\em 2017 IEEE International Conference on Computer Vision
  (ICCV)}, pages 2980--2988, 2017.

\bibitem{resnet}
Kaiming He, Xiangyu Zhang, Shaoqing Ren, and Jian Sun.
\newblock Deep residual learning for image recognition, 2015.

\bibitem{thirdplace}
Eugene Khvedchenya.
\newblock xview2\_third\_place.
\newblock \url{https://github.com/DIUx-xView/xView2_third_place}, 2020.

\bibitem{kwcoco}
Jon~Crall Kitware.
\newblock Kwcoco.
\newblock \url{https://github.com/Kitware/kwcoco}, 2021.

\bibitem{fifthplace}
Jamyoung Koo.
\newblock xview2\_fifth\_place.
\newblock \url{https://github.com/DIUx-xView/xView2_fifth_place}, 2020.

\bibitem{lin2017focal}
Tsung-Yi Lin, Priya Goyal, Ross Girshick, Kaiming He, and Piotr Doll{\'a}r.
\newblock Focal loss for dense object detection.
\newblock In {\em Proceedings of the IEEE international conference on computer
  vision}, pages 2980--2988, 2017.

\bibitem{coco}
Tsung-Yi Lin, Michael Maire, Serge Belongie, Lubomir Bourdev, Ross Girshick,
  James Hays, Pietro Perona, Deva Ramanan, C.~Lawrence Zitnick, and Piotr
  Dollár.
\newblock Microsoft coco: Common objects in context, 2014.

\bibitem{maxar_open_data}
Maxar.
\newblock Maxar open data program.
\newblock \url{https://www.maxar.com/open-data}, 2022.

\bibitem{milletari2016v}
Fausto Milletari, Nassir Navab, and Seyed-Ahmad Ahmadi.
\newblock V-net: Fully convolutional neural networks for volumetric medical
  image segmentation.
\newblock In {\em 2016 fourth international conference on 3D vision (3DV)},
  pages 565--571. IEEE, 2016.

\bibitem{poisson_blending}
Patrick P\'{e}rez, Michel Gangnet, and Andrew Blake.
\newblock Poisson image editing.
\newblock {\em ACM Trans. Graph.}, 22(3):313–318, jul 2003.

\bibitem{secondplace}
Selim Seferbekov.
\newblock xview2\_second\_place.
\newblock \url{https://github.com/DIUx-xView/xView2_second_place}, 2020.

\bibitem{efficientnet}
Mingxing Tan and Quoc~V. Le.
\newblock Efficientnet: Rethinking model scaling for convolutional neural
  networks, 2019.

\bibitem{fourthplace}
Zhuo Zheng.
\newblock xview2\_fourth\_place.
\newblock \url{https://github.com/DIUx-xView/xView2_fourth_place}, 2020.

\bibitem{zheng2021building}
Zhuo Zheng, Yanfei Zhong, Junjue Wang, Ailong Ma, and Liangpei Zhang.
\newblock Building damage assessment for rapid disaster response with a deep
  object-based semantic change detection framework: From natural disasters to
  man-made disasters.
\newblock {\em Remote Sensing of Environment}, 265:112636, 2021.

\end{thebibliography}


\begin{thebibliography}{1}\itemsep=-1pt

\bibitem{girshick2015fast}
Ross Girshick.
\newblock Fast r-cnn.
\newblock In {\em Proceedings of the IEEE international conference on computer
  vision}, pages 1440--1448, 2015.

\bibitem{lin2017focal}
Tsung-Yi Lin, Priya Goyal, Ross Girshick, Kaiming He, and Piotr Doll{\'a}r.
\newblock Focal loss for dense object detection.
\newblock In {\em Proceedings of the IEEE international conference on computer
  vision}, pages 2980--2988, 2017.

\bibitem{liu2020multiscale}
Yuanyuan Liu, Dingyuan Chen, Ailong Ma, Yanfei Zhong, Fang Fang, and Kai Xu.
\newblock Multiscale u-shaped cnn building instance extraction framework with
  edge constraint for high-spatial-resolution remote sensing imagery.
\newblock {\em IEEE Transactions on Geoscience and Remote Sensing},
  59(7):6106--6120, 2020.

\end{thebibliography}
}

\end{document}


\title{Supplemental to xFBD: Focused Building Damage Dataset and Analysis}

\maketitle

\section{Example of Importance of Context in xBD}
\begin{figure}
     \centering
     \begin{subfigure}[b]{0.49\linewidth}
         \centering
         \includegraphics[width=\linewidth]{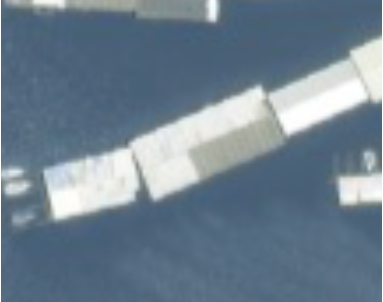}
         \caption{This building is only correctly classifiable as ``destroyed'' by looking at the context: surrounding floodwater. The building itself looks fine.}
         \label{fig:building_only_classify_by_context}
     \end{subfigure}
     \hfill
     \begin{subfigure}[b]{0.49\linewidth}
         \centering
         \includegraphics[width=\linewidth]{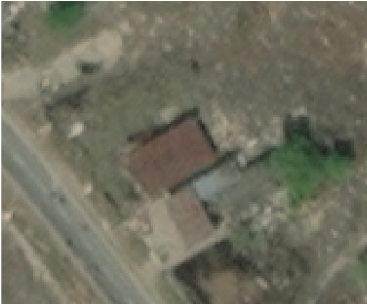}
         \caption{Here, looking at the context is a bad prior for classifying this building. It appears to be undamaged, despite being surrounded by wreckage.}
         \label{fig:building_context_is_bad_prior}
     \end{subfigure}

    \caption{Examples of the changing importance of context in different areas of the xBD dataset.}
    \label{fig:building_context}
\end{figure}

Figure \ref{fig:building_context} shows two examples of buildings which might be difficult for an xView2 network to make an accurate decision about.
The first (Figure \ref{fig:building_only_classify_by_context}) can only be correctly classified by looking at the surrounding area, while the second (Figure \ref{fig:building_context_is_bad_prior}) can only be correctly classified if the network focuses on just how the building looks and not the surrounding area.

\section{Samples from xFBD Dataset}
\begin{figure*}
    \centering
    \includegraphics[width=0.82\linewidth]{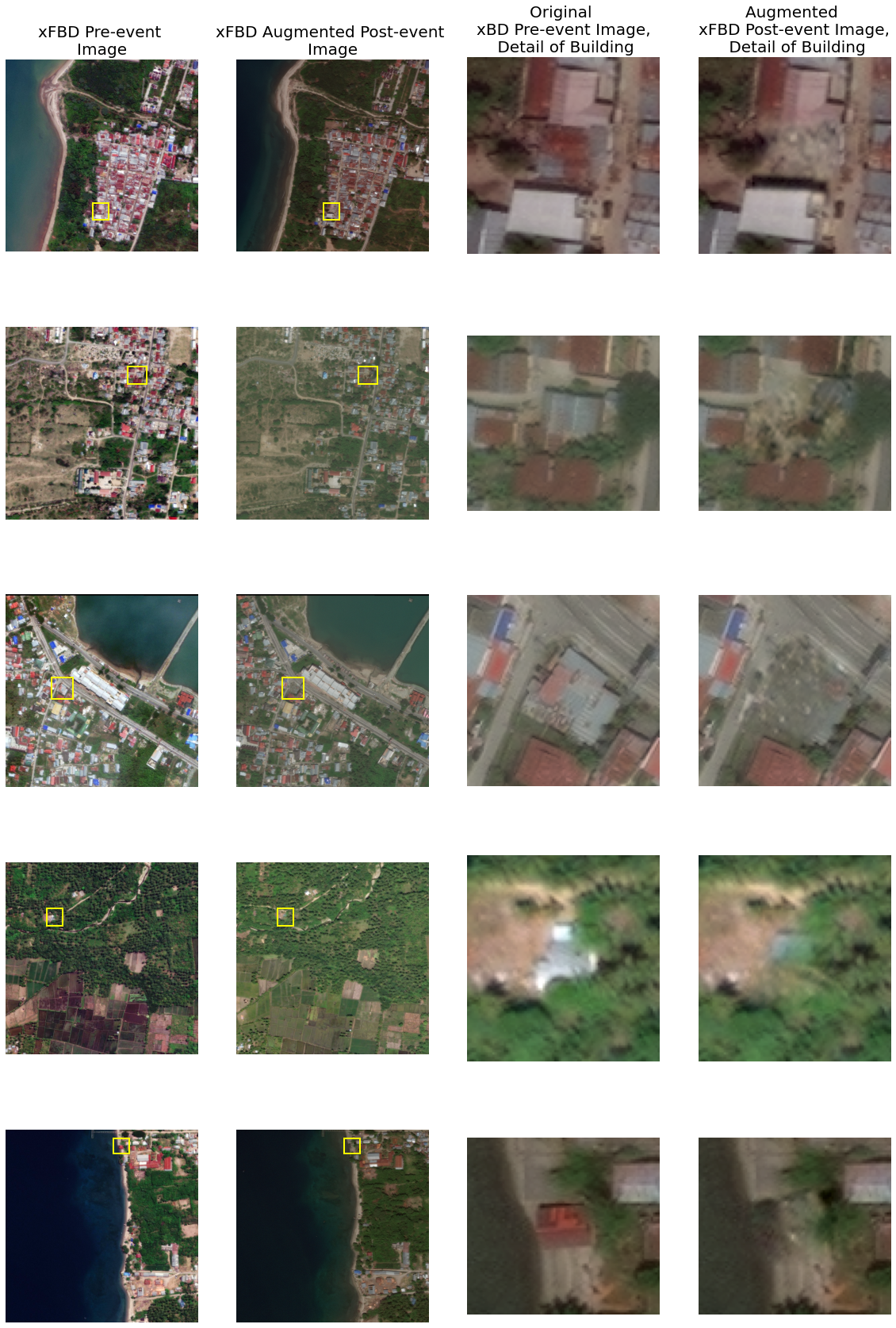}
    \caption{Samples from the xFBD dataset, with a detail on where the single damaged post-event building is located (last column).}
    \label{fig:xfbd_examples}
\end{figure*}

In Figure~\ref{fig:xfbd_examples} we show five examples from the xFBD dataset. 
In the first column is the secondary pre-event image pulled from Maxar's Open Data Program, used as the pre-event input to a network.
In the second column is the augmented post-event image. 
This is the original xBD pre-event image, with a single damaged post-event building blended in.
This building is indicated by the yellow box.
In the third column is a detail of what the xBD pre-event image looked like, before a post-event building was blended in.
In the fourth column is the same building detail, but with the blending completed. 
This last column is thus a small portion of the image from the second column, zoomed in for clarity.

\ifarxiv

\else
\section{Included Video}
The included video, \verb|blend_heatmap.gif|, is a moving version of Figure~\ref{fig:blend_unchanging_heatmap} in the main body of the paper.
It shows the same original xBD pre-event image with different buildings being blended in, and the network damage output for these different blends.
It is interesting to note that as the building being blended changes, the network does not generally change its output. 
Large buildings do seem to trigger a change.
\fi

\begin{table*}[t]
    \begin{center}
        \begin{tabular}{l|l|ccccc}
            \hline
             & Method & Localization & Destroyed & Major Damage & Minor Damage & No Damage \\
            \hline
            \parbox[t]{2mm}{\multirow{6}{*}{\rotatebox[origin=c]{90}{Pixel-Level}}} & 4PS & \textbf{0.860} & \textbf{0.822} & 0.719 & 0.508 & \textbf{0.922} \\
            & 4PS-Edge & 0.849 & 0.821 & \textbf{0.760} & 0.573 & 0.921 \\
            & 4PS-Focal & 0.857 & 0.794 & 0.694 & 0.513 & 0.907 \\
            & 4PS-Focal-v1 & 0.832 & 0.799 & 0.728 & 0.559 & 0.916 \\
            & 4PS-Focal-v2 & 0.742 & 0.782 & 0.724 & 0.318 & 0.918 \\
            & 4PS-Contour & 0.849 & 0.787 & 0.703 & \textbf{0.569} & 0.908\\
            & 4PS-Localization & 0.795  & 0.203 & 0.000 & 0.000 & 0.808 \\
            & Mask-RCNN &0.699 & 0.605 & 0.429& 0.277&0.675 \\
            \hline
            \parbox[t]{2mm}{\multirow{6}{*}{\rotatebox[origin=c]{90}{Object-Level}}} & 4PS &  0.440 & 0.420 & 0.340 & 0.240 & 0.420 \\
            & 4PS-Edge & 0.410 & 0.370 & 0.320 & 0.240 & 0.370 \\
            & 4PS-Focal & 0.440 & \textbf{0.430} & \textbf{0.360} & 0.210 & 0.400 \\
            & 4PS-Focal-v1 & 0.350 & 0.330 & 0.270 & 0.210 & 0.310 \\
            & 4PS-Focal-v2 & 0.190 & 0.190 & 0.150 & 0.070 & 0.170 \\
            & 4PS-Contour & 0.410 & 0.350 & 0.310 & \textbf{0.240} & 0.360\\
            & 4PS-Localization & 0.280 & 0.040 & 0.000 & 0.000 & 0.230 \\
            & Mask-RCNN & \textbf{0.470} & 0.280 & 0.300 &0.180 & \textbf{0.450} \\
            \hline
        \end{tabular}
    \end{center}
    
    \caption{New Baselines proposed, tested on the original xBD holdout set.}
    \label{table:xbd_supplement_baselines}
\end{table*}

\begin{table*}
\begin{center}
\begin{tabular}{l|l|ccccc}
            \hline
             & Method & Localization & Destroyed & Major Damage & Minor Damage & No Damage \\
            \hline
            \parbox[t]{2mm}{\multirow{6}{*}{\rotatebox[origin=c]{90}{Pixel-Level}}} & 4PS & \textbf{0.758} & \textbf{0.066} & \textbf{0.016} & n/a & 0.907  \\
            & 4PS-Edge & 0.739 & 0.045 & 0.015 & n/a & 0.892 \\
            & 4PS-Focal & 0.742 & 0.042 & 0.000 & n/a &  0.873\\
            & 4PS-Contour & 0.741 & 0.060 & 0.000 & n/a & 0.887\\
            & 4PS-Localization & 0.732 & 0.000 & 0.000 & n/a & 0.899  \\
            & Mask-RCNN & 0.494 & 0.013& 0.011& n/a & 0.522\\
            \hline
            \parbox[t]{2mm}{\multirow{6}{*}{\rotatebox[origin=c]{90}{Object-Level}}} & 4PS & 0.040 & 0.000 & 0.000 & n/a & 0.040  \\
            & 4PS-Edge & 0.040 & 0.000 & 0.000 & n/a & 0.040 \\
            & 4PS-Focal & 0.040 & \textbf{0.010} & 0.000 & n/a & 0.010  \\
            & 4PS-Focal-v1 & 0.040 & 0.000 & 0.000 & n/a & 0.040 \\
            & 4PS-Focal-v2 & 0.020 & 0.000 & 0.000 & n/a & 0.020 \\
            & 4PS-Contour & 0.050 & 0.000 & 0.000 & n/a & 0.050\\
            & 4PS-Localization & 0.030 & 0.000 & 0.000 & n/a & 0.030 \\
            & Mask-RCNN & \textbf{0.210} & 0.000 & 0.000 & n/a & \textbf{0.200} \\
            \hline
        \end{tabular}
\end{center}

\caption{New baselines proposed, tested on the single-building-paste dataset (xFBD)}
\label{table:xfbd_supplement_baselines}
\end{table*}

\section{Additional Baselines}

We show several additional baselines we implemented which were not as successful as others, in the interest of reducing the time other researchers may spend on exploring this space. The results are shown in Tables \ref{table:xbd_supplement_baselines} and \ref{table:xfbd_supplement_baselines}.

\begin{enumerate}[leftmargin=*]
    \item \textbf{4PS-Edge} consists of the classification loss and the Edge Constraints Loss (ECL) \cite{liu2020multiscale} together. According to the detected instance positions, 4PS-Edge is proposed to extract each precise mask and suppress overfitting by using a hybrid loss function, where the goal is to constrain the differences between the boundaries of the predicted instance mask and the ground truth. This loss function takes the form:
    \begin{align}
        L_{Edge}(\mathbf{y},\mathbf{\hat{y}}) = L_{cls}(\mathbf{y},\mathbf{\hat{y}})+\lambda_1\cdot L_{ecl}(\mathbf{y},\mathbf{\hat{y}})
    \end{align}
    where $\mathbf{y},\mathbf{\hat{y}}$ represent the ground truth and predicted values, respectively.
    $\lambda_1=1$ and $L_{cls}(\mathbf{y},\mathbf{\hat{y}})$ is the the normal cross-entropy loss which is $L_{cls}(\mathbf{y},\mathbf{\hat{y}})=-\sum_i^n\big(y_i\log\hat{y}_i+(1-y_i)(1-\log\hat{y}_i)\big)$ where $n$ is the total pixel number of an image. The ECL can be defined as :
    \begin{align}
        L_{ecl}(\mathbf{y},\mathbf{\hat{y}}) = \frac{\big( \mathbf{1}^T(\mathbf{y}-\mathbf{\hat{y}}) \big)^2}{2}
    \end{align}

    
    \item \textbf{4PS-Focal-v1} \cite{lin2017focal} shares the same loss function stated in Eq.(\ref{eq:focal}). In contrast to the proposed \textit{4PS-Focal}, $h(\cdot)$ is a piece-wise function, where
    \begin{align}
    \label{eq:piecewise_func}
        h(y_i,\hat{y}_i) = 
            \begin{cases}
                  \hat{y}_i \quad &\text{if} \quad y_i=1 \\
                  1-\hat{y}_i \quad &\text{if} \quad \text{otherwise} \\
            \end{cases}
    \end{align}
    Here, when $\gamma=0$, the loss function is equivalent to $L_{cls}$ introduced in \textit{4PS-Edge}. We set $\gamma=2$ for our experiments.
    
    \item \textbf{4PS-Focal-v2}. An alternate instantiation of \textit{4PS-Focal} is proposed, in which the modulating factor is removed. This loss is defined as:
    \begin{align}
        L_{Focal\_v2}(\mathbf{y},\mathbf{\hat{y}}) = \frac{-\log(h(\mathbf{y},\gamma\mathbf{y\odot\hat{y}}+\beta))}{\gamma}
    \end{align}
    where $h(\cdot)$ is the piecewise function defined in Eq.(\ref{eq:piecewise_func}). $\gamma$ and $\beta$ are parameters controlling the steepness and shift of the loss curve. In practice, we set $\gamma=2$ and $\beta=1$.

    \item \textbf{4PS-Localization} combines the classification loss $L_{cls}$ in 4PS-Edge and a Smooth L1 loss introduced in \cite{girshick2015fast}.
    \begin{align}
        L_{loc}(\mathbf{y,\hat{y}})&=L_{cls}(\mathbf{y,\hat{y}})\\
                                    &+\lambda_4 \cdot [\mathbf{y}\succcurlyeq\mathbf{1}]^T \text{smooth}_{L1}(\mathbf{y-\hat{y}}) \nonumber
    \end{align}
    where the Iverson bracket indicator function $[y_i\geq 1 ]$ evaluates to 1 when $y_i\geq 1$ and 0 otherwise. For $L_{smooth_{L1}}$, it takes the form
    \begin{align}
        \text{smooth}_{L1}(x) = 
            \begin{cases}
                  0.5x^2 \quad &\text{if} \quad |x|<1 \\
                  |x|-0.5 \quad &\text{if} \quad \text{otherwise} \\
            \end{cases}
    \end{align}
    Empirically, we set $\lambda_4=1$.

\end{enumerate}

{\small
\bibliographystyle{ieee_fullname}
\bibliography{egbib}
}